# Multi-Vehicle Routing Problems with Soft Time Windows: A Multi-Agent Reinforcement Learning Approach

Ke Zhang[a], Fang He[b], Zhengchao Zhang[a], Xi Lin[a], Meng Li[a]*

[a]*Department of Civil Engineering, Tsinghua University, Beijing 100084, P.R. China*

[b]*Department of Industrial Engineering, Tsinghua University, Beijing 100084, P.R. China*

**Abstract**

Multi-vehicle routing problem with soft time windows (MVRPSTW) is an indispensable constituent in urban logistics distribution systems. Over the past decade, numerous methods for MVRPSTW have been proposed, but most are based on heuristic rules that require a large amount of computation time. With the current rapid increase of logistics demands, traditional methods incur the dilemma between computational efficiency and solution quality. To efficiently solve the problem, we propose a novel reinforcement learning algorithm called the Multi-Agent Attention Model that can solve routing problem instantly benefit from lengthy offline training. Specifically, the vehicle routing problem is regarded as a vehicle tour generation process, and an encoder-decoder framework with attention layers is proposed to generate tours of multiple vehicles iteratively. Furthermore, a multi-agent reinforcement learning method with an unsupervised auxiliary network is developed for the model training. By evaluated on four synthetic networks with different scales, the results demonstrate that the proposed method consistently outperforms Google OR-Tools and traditional methods with little computation time. In addition, we validate the robustness of the well-trained model by varying the number of customers and the capacities of vehicles.

**Keywords:** Reinforcement learning; Vehicle routing problem; Attention mechanism; Computational efficiency; Multi-agent

---

* Corresponding author.

# 1 INTRODUCTION

City logistics has become an active research field in both industrial and academic environments due to the rapid development of urbanization over the past few decades. Currently, the logistics demand is growing fast worldwide. In China, the express delivery market recorded over 50 billion orders in 2018 (an increase of 26.6% from last year) (State Post Bureau of China, 2019). In Germany and the US, it is expected that the parcel delivery market will be growing annually at 7% to 10% in mature markets, and delivery volumes in 2025 could reach approximately 5 billion and 25 billion parcels, respectively (Joerss et al., 2016). The rapid growth of the logistics industry has brought new challenges to the operation of large-scale systems for serving massive requests within a short period of time.

The vehicle routing problem (VRP) is one of the most important topics in urban logistics; in this problem, many customers are to be served by a fleet of vehicles that has limited capacity, and the fleet manager aims to minimize the service cost under some service constraints (Toth et al., 2002; Kumar et al., 2012). In real applications, VRPs always involve a fleet of vehicles set off from a depot to serve many customers with various demands and within specific time window constraints. The given time windows can sometimes be violated, but with associated penalties (e.g., compensation to customers, customers' negative evaluations, and so on); such a constraint is often called a *soft time window* constraint. As a result, the problem to be investigated is called multiple vehicle routing problem with soft time windows (MVRPSTW) (Lau et al., 2003).

MVRPSTW is a well-known NP-hard problem that has drawn much attention from many researchers for many decades. Although many heuristic algorithms have been proposed to address the MVRPSTW, such as the iterated local search (Ibaraki et al., 2008), genetic algorithm (Louis et al., 1999; Wang et al., 2008), tabu search method (Lim et al., 2004) and adaptive large neighborhood search (Tas et al., 2014), the ability to provide fast and reliable solutions is still a challenging task. Moreover, the heuristic algorithms have difficulty solving the scenario in a short span of time with a proliferation of demands, which makes them unable to support large-scale applications. On the other hand, the canonical mixed-integer programming (MIP) method is hardly a good option for this problem due to the existence of soft time windows, i.e., the mathematical structure of the penalty term causes nonlinearity, and resolving such a nonlinearity requires introducing a very large number of binary variables.

Recently, machine-learning-based methods have become increasingly prominent in numerous research fields owing to their excellent learning ability (Angra et al., 2017; Zhang et al., 2019). Deep reinforcement learning (DRL) especially shows great power in solving complex time-dependent operation problems because it benefits from the parameter training process, which learns the solution space characteristics and establishes a parameterized computation graph to emulate the constraints of the original problems (Arulkumaran et al., 2017). In detail, DRL utilizes a self-driven learning procedure that only requires a reward calculation based on the generated outputs. Once a generated sequence is feasible and its reward is derived, the desired meta algorithm can be learned (Bello et al., 2016). This approach provides a general framework for optimizing decisions in dynamic environments, which can help to solve combination optimization problems. Significant attention has also been attracted to model the VRP utilizing the DRL framework. Several highly related research studies are listed as follows:

Vinyals et al. (2015) proposes a pointer network to solve the traveling salesman problem (TSP) by generating a permutation of the input routes with the attention mechanism. Bello et al. (2016) introduces neural combinatorial optimization, a framework to tackle TSP with reinforcement learning and neural networks. Experiments demonstrate that their method nearly achieves the optimal results on Euclidean graphs with up to 100 customers. Nazari et al. (2018) applies a policy gradient algorithm (Silver et al., 2014) to solve VRP, which consists of a recurrent neural network (RNN) decoder coupled with an attention mechanism. After training, the model can find near-optimal solutions for VRP with split deliveries, which is also available for stochastic variant instances of similar size in real time without retraining. Khalil et al. (2017) combines the deep Q-learning (DQN) algorithm (Mnih et al., 2015) and graph embedding (Dai et al., 2016) to address the TSP problem. They trained the model to construct a solution in which the nodes are inserted into a partial tour, and the action is determined by the output of a graph embedding network that captures the current state of the solution. Kool et al. (2018) propose an encoder-decoder framework with multi-head attention layers (Vaswani et al., 2017) to solve VRP and use a reinforce gradient estimator with a simple baseline (Williams, 1992) based on a deterministic greedy rollout to train the model. The training strategy is more efficient than the manner of using a value function.

To summarize, these pioneering studies have produced fruitful results in the field of single vehicle dispatching problems, mainly by applying reinforcement learning on combination optimization problems. Those research studies make use of the generalization capability of artificial intelligence to develop vehicle routes with satisfactory performance. However, few studies have attempted to employ machine-learning-based methods to solve VRPs with multiple vehicles and soft time windows; tackling this type of problem requires the resolution of the following difficulties: i) the existence of multiple vehicles requires suitable methodological development to handle multi-agent coordination in a time-dependent environment; ii) the settings of the soft time windows must be appropriately incorporated into the methodological framework; iii) the method should be able to solve problems with considerable scale-up; and iv) the method should reach a high-quality solution with acceptable computational effort.

In this work, we propose a novel reinforcement learning architecture called the multi-agent attention model (MAAM), which is based on recent advances in deep learning techniques, to efficiently solve MVRPSTW. First, we construct an encoder-decoder framework with multi-head attention layers to iteratively develop routes for vehicles in the system, which utilizes a deep reinforcement learning strategy to determine the model parameters. Most notably, a new multi-agent reinforcement learning method based on multiple vehicle context embedding is proposed to handle the interactions among the vehicles and customers. After lengthy offline training, the model can be deployed instantly without further training for new problems. In contrast to solving a complex problem without an explicit analytical form, this method is quite appealing since it only requires a verifier to find feasible solutions and the corresponding rewards which demonstrate how well the model performs. Our numerical experiments indicate that our framework performs significantly better than Google OR-Tools and the well-known classical heuristics that designed for the MVRPSTW. Additionally, we validate the robustness of the proposed model with extensive case studies, which indicates that the new instances with different numbers of customers and vehicle capacities do not require retraining to obtain desirable solutions.

The remainder of this paper is organized as follows. Section 2 states some preliminaries and defines the problem in details. Section 3 describes the overall architecture and mathematical formulations of the proposed MAAM. In Section 4, the model performance is evaluated by performing comprehensive case studies. Furthermore, we analyze the parameter settings and discuss the robustness of a well-trained model with varying customer numbers and vehicle capacities. Finally, we conclude the paper and outline the future work in Section 5.

## 2 PROBLEM DEFINITION

We first describe the notations of the variables used herein. The road network can be regarded as a fully connected graph with a randomly generated depot and customers in the Euclidean plane. Let $G(\mathfrak{V})$ be a connected graph, where $\mathfrak{V} = \{v_0, v_1, \dots, v_N\}$. A problem instance is denoted by a tuple $s = (\boldsymbol{v}, \boldsymbol{d}, \boldsymbol{e}, \boldsymbol{l}, \boldsymbol{\alpha}, \boldsymbol{\beta})$, where $v_0$ is the depot with coordinate $\boldsymbol{x_0}$, $v_i(i \neq 0)$ denotes a customer with coordinate $\boldsymbol{x_i}$, the demand is $d_i$, with time windows $(e_i, l_i)$, and early and late penalty coefficients $\alpha_i$, $\beta_i$. Given a fleet of identical vehicles (each with capacity $Q$) and a problem instance $s = (\boldsymbol{v}, \boldsymbol{d}, \boldsymbol{e}, \boldsymbol{l}, \boldsymbol{\alpha}, \boldsymbol{\beta})$, the decision maker assigns $M$ capacitated vehicles to serve all customer requests, and the goal of the problem is to find a set of minimum cost vertex-disjoint routes $\boldsymbol{r[m]}$ $(m = 1,2,\dots,M)$ for each vehicle, while both starting and ending at depot $v_0$. Under this circumstance, each customer $v_i$ is served only once by one of the vehicles within its time windows. Figure 1 gives an overview of the problem scenario.

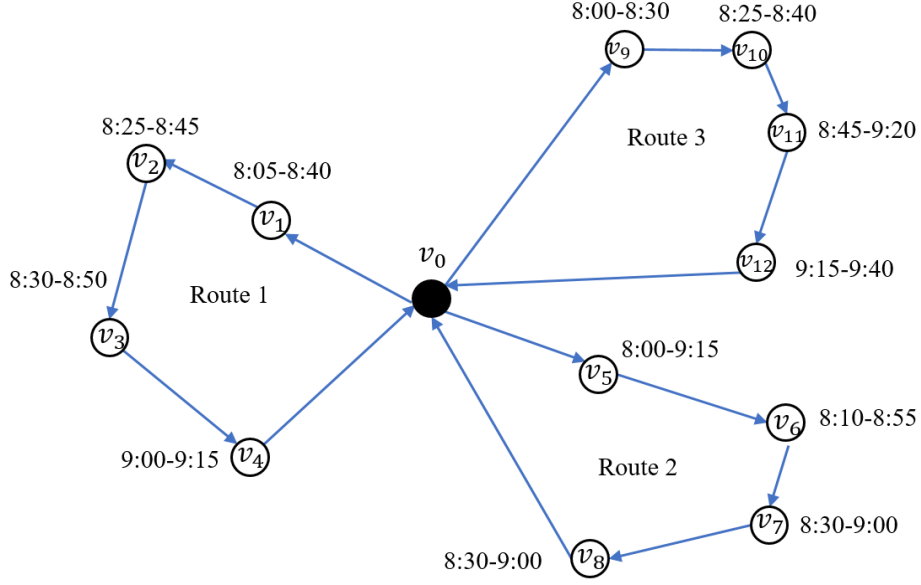

Figure 1: Illustration of a VRP

Table 1 summarizes the notations adopted to define the problem. Since each vehicle is dedicated to a unique route, a total number of $M$ routes will be generated, and they only connect to each other at the depot. All distances are represented by Euclidean distances in the plane, and the speeds of all vehicles are assumed to be identical (i.e., it takes one unit of time to travel one unit of distance). The remaining capacity $\hat{d}_{m,t}$ of the $m^{th}$ vehicle at timestep $t$ must be greater than zero, which means that no vehicles can be overloaded. The problem is to find a solution $r[1, M] = (r[1], r[2], \ldots, r[M])$ with minimal total cost, which is defined as follows:

$$\text{Cost}(r[1, M]) = d_{sum}(r[1, M]) + p_{sum}(r[1, M]), \quad (1)$$

where $d_{sum}(r[1, M]) = \sum_{m=1}^{M} \sum_{b=1}^{|r[m]|-1} ||x_{r[m][b]}, x_{r[m][b+1]}||_2$ is the total travel cost of all vehicles, and $p_{sum}(r[1, M]) = \sum_{m=1}^{M} \sum_{i=1}^{N} \left[ I_{(e_i > \tilde{t}_i)} * \alpha_i * (e_i - \tilde{t}_i) + I_{(\tilde{t}_i > l_i)} * \beta_i * (\tilde{t}_i - l_i) \right]$ denotes the total penalty for the time window constraints, where $\tilde{t}_i$ represents the time when a vehicle serves customer $i$; the penalty is a piecewise linear function of the arrival time, as illustrated in Figure 2. Arriving earlier at any customer is considered to be an early arrival penalty in the MVRPSTW, which is usually much smaller than the late arrival penalty.

Table 1: Nomenclature

| Symbol | Definition |
|---|---|
| $x_i$ | Coordinate of customer $i$ |
| $d_i$ | Demand of customer $i$ |
| $M$ | Vehicle number |
| $r[1,m]$ | Tours of $1^{th},...,m^{th}$ vehicles |
| $r[m][t]$ | Customer served by $m^{th}$ vehicle at timestep $t$ |
| $c[t]$ | The customers have been served at timestep $t$ |
| $\tilde{t}_i$ | Total travel time for one vehicle arriving at customer $i$ |
| $\hat{d}_{m,t}$ | Remaining capacity of $m^{th}$ vehicle at timestep $t$. |
| $[e_i, l_i]$ | Time window of customer $i$ |
| $\alpha_i$ | Early penalty coefficients for customer $i$ |
| $\beta_i$ | Late penalty coefficients for customer $i$ |

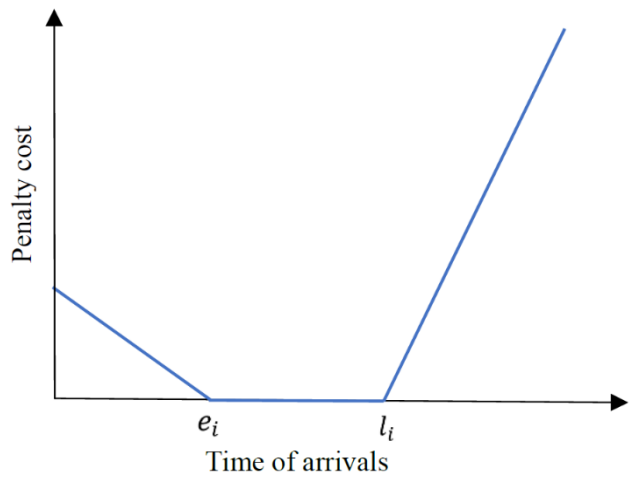

Figure 2: Penalty function for the MVRPSTW

# 3 MULTI-AGENT ATTENTION MODEL

In this subsection, we propose the novel multi-agent attention model (MAAM), which is essentially an attention-based encoder-decoder structure. Following the approach of deep reinforcement learning (DRL), we regard the MVRPSTW problem as a dynamic route generation problem, which treats the solution as a sequence of decisions. The details of the proposed model are presented as follows.

## 3.1 Overview of MAAM

As shown in Figure 3, the encoder produces the embedding of the depot and all customers. Then, the decoder incorporates the outputs of the encoder, the mask matrix for the constraints and the context embedding as inputs, and consequently, it produces a sequence $r[1, M]$ of input customers, one customer for one vehicle at a timestep. When a sub-tour has been constructed, the problem at that time is to find a path from the last customer for each vehicle's sub-tour through all unvisited customers to the depot. At that time, the requests of other customers already visited are irrelevant to the decision-making. Our model defines a stochastic policy $p(r[1, M]|s)$ for selecting a solution $r[1, M]$ given a problem instance $s$, which is defined in Section 2. It is factorized and parameterized by $\theta$ as follows:

$$p_\theta(r[1, M]|s) = \prod_{t=1}^{n} p_\theta(r[\mathcal{M}(t)][t]|s, c[t-1]) \qquad (2)$$

In Eq.(2), $\mathcal{M}(t) = t \pmod{M}$, which means the remainder of timestep $t$ divided by the vehicle number M. $p_\theta(r[\mathcal{M}(t)][t]|s, c[t-1])$ represents the probability of choosing customer $r[\mathcal{M}(t)][t]$ at timestep $t$ for vehicle $\mathcal{M}(t)$ given the problem instance $s$ and given the customers that have been served at timestep $t-1$. Here, $n$ represents the timestep at which all customers have been served. $p_\theta(r[1, M]|s)$ represents the stochastic policy for selecting a solution, and it also plays a critical role in the training method.

Instead of training a policy for every problem instance separately, this method can perform well on any problem instance generated from the given distributions randomly.

Therefore, we can apply the well-trained model to generate a high-quality solution as a sequence of consecutive actions without retraining for every new problem in a short time period.

The specific components of the MAAM are demonstrated as follows.

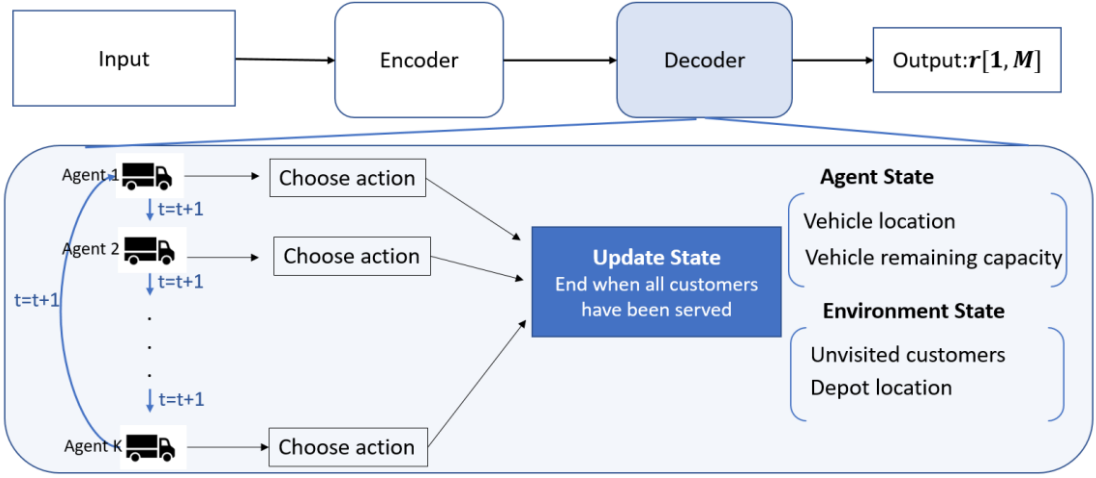

Figure 3: The Multi-Agent Attention Model

### 3.2 Encoder Framework

The encoder framework is similar to that of Kool et al. (2018), which stays invariant to the input order. First, it computes the initial customer embedding $\boldsymbol{h}_i^{(0)}$ through a learnable linear projection with parameters $\boldsymbol{W^1}$ and $\boldsymbol{b^1}$:

$$\boldsymbol{h}_i^{(0)} = \boldsymbol{W^1} \cdot [\boldsymbol{x_i}, d_i, e_i, l_i] + \boldsymbol{b^1} \tag{3}$$

where $\boldsymbol{x_i}, d_i, e_i, l_i$ are defined in Section 2, and the operator $[\,\cdot\,,\,\cdot\,]$ concatenates two tensors along the same dimensions.

The embedding is updated using multiple attention layers. Each attention layer carries out a multi-head attention and a feed-forward operation. The attention mechanism can be interpreted as a weighted message passing algorithm between customers in a graph (Vaswani et al., 2017). The weight of the message that the customer receives from other customers depends on the compatibility of its query with the key of other customers. Formally, a single attention function is given by

$$u_{i,j} = \frac{q_i^T k_j}{\sqrt{d_k}} \quad (4)$$

$$h'_i = \text{softmax}(u_{i,j}) v_j = \sum_j \frac{e^{u_{i,j}}}{\sum_{j'} e^{u_{i,j'}}} v_j \quad (5)$$

where $k_i = W^K h_i^{(0)}$, $v_i = W^V h_i^{(0)}$ and $q_i = W^Q h_i^{(0)}$ are the key, value and query for each customer by projecting the embedding $h_i^{(0)}$. Here, $u_{i,j}$ calculates the compatibility of the query $q_i$ of customer $i$ with the key $k_j$ of customer $j$ in the method of scaled dot-product. Here, $d_k$ is the vertical dimension of $h_i^{(0)}$, which is used to scale the dot products and avoid an overflow of numerical calculations. Additionally, $h'_i$ is the output of the attention function.

Furthermore, multi-head self-attention is employed for feature augmentation, which allows the model to attend to information jointly from different representation subspaces at different positions (Vaswani et al., 2017). In detail, we compute the attention value $Z$ times with different parameters, with each result represented by $h'_{iz}$ for $z \in \{1,2,\dots,Z\}$. The final multi-head attention value for customer $i$ is a function of $h_1^{(0)}, h_2^{(0)}, \dots, h_n^{(0)}$:

$$\mathcal{F}_i\left(h_1^{(0)}, h_2^{(0)}, \dots, h_n^{(0)}\right) = \sum_{z=1}^{Z} W_Z^O h'_{iz} \quad (6)$$

The remainder of the attention layer is a feed-forward operation $F$ with skip-connections (He et al., 2016):

$$\hat{h}_i = h_i^{(0)} + \mathcal{F}_i\left(h_1^{(0)}, h_2^{(0)}, \dots, h_n^{(0)}\right) \quad (7)$$

$$h_i^{(1)} = \hat{h}_i + \varphi(\hat{h}_i) \quad (8)$$

where the operation $\varphi$ is defined as

$$\varphi(\hat{h}_i) = W_1^f \text{ReLu}\left(W_2^f \hat{h}_i + b_0^f\right) + b_1^f \quad (9)$$

We compute Eq. (6)-(8) $\lambda$ times to acquire $\{h_i^{(\lambda)}, i = 1, \ldots, n\}$. Finally, the encoder computes an aggregated embedding of all customers as the mean of the final output layer:

$$\bar{h}^{(N)} = \frac{1}{n} \sum_{i=1}^{N} h_i^{(\lambda)} \tag{10}$$

**3.3 Decoder Framework**

In the decoder part, we design the state, action space, and reward in an explicit manner, and we model each agent by deep neural networks. We regard vehicles as agents that perceive the state from the environment and each other. Then, they decide a sequential action set based on the knowledge obtained through this perception. The action taken affects the environment and, consequently, changes the state in which the agent is. Every agent within the DRL system has a goal state that must be achieved. The goal of the agent is to maximize such long-term rewards by learning a good policy, which is a mapping from perceived states to actions (Arel et al., 2010). To approach the problem with reinforcement learning, the following subsections provide the principles and fundamental components of reinforcement learning for the route generation policy, including the environment and its states, the action set, reward function and algorithm.

At timestep $t$, the decoder outputs the next customer to serve based on the embedding from the encoder and the previous outputs $c[t-1]$. The main objective of the agent is to select a sequence of actions up to the goal state, which maximizes the reinforcement accumulated over time. Thus, a decision policy is generated, which is characterized by the mapping of states and actions. The decoder process will be end when all of the customers have been served.

*(1) State*

The global state can be divided into the environment state and agent state. The environment state contains the final embedding of the customers $\bar{h}^{(N)}$ and the already visited customers. The agent state consists of the current vehicle location and its remaining capacity. At each decoding timestep, the vehicle chooses the customers to visit in the next step. After visiting customer $i$, the remaining capacity $\hat{d}_{m,t}$ of vehicle $m$ is updated as follows:

$$\hat{d}_{m,t} = \max(0, \hat{d}_{m,t-1} - d_i) \tag{11}$$

To utilize the information of the state, we define multiple vehicles context embedding $\boldsymbol{h}_t^{(c)}$ for the decoder at timestep $t$, which comes from the encoder and the vehicle output up to timestep $t$:

$$\boldsymbol{h}_t^{(c)} = \left[\bar{\boldsymbol{h}}^{(N)}; \boldsymbol{h}_{r_{t-1},1}^{(N)}; \hat{d}_{1,t}; \boldsymbol{h}_{r_{t-1},2}^{(N)}; \hat{d}_{2,t}; \dots; \boldsymbol{h}_{r_{t-1},M}^{(N)}; \hat{d}_{M,t}\right] \tag{12}$$

*(2) Action*

The action for each vehicle represents the choice of the next customer to be visited at timestep $t$. First, we compute a new multiple vehicles context embedding $\boldsymbol{h}_t^{(c)\prime}$ using the multi-head attention mechanism:

$$\boldsymbol{h}_t^{(c)\prime} = \text{MHA}\left(\boldsymbol{h}_t^{(c)}\right) \tag{13}$$

Then, we compute the compatibility of the query $\boldsymbol{q}_{(c)}$ with all customers:

$$\boldsymbol{q}_{(c)} = \boldsymbol{W}^Q \boldsymbol{h}_t^{(c)\prime} \tag{14}$$

$$\boldsymbol{k}_i = \boldsymbol{W}^K \boldsymbol{h}_i^{(\lambda)} \tag{15}$$

$$u_{i,m,t} = \tanh\left(\frac{\boldsymbol{q}_{(c)}^T \boldsymbol{k}_i}{\sqrt{d_k}}\right) \tag{16}$$

Similar to Bello et al. (2016), the decoder observes a mask to know which customers have been visited. We mask the (set $u_{i,m,t} = -\infty$) customers, which have been visited before timestep $t$ or whose demand exceeds the vehicle's remaining capacity.

Finally, we regard these compatibilities as unnormalized log probabilities, and we compute the probability of choosing customer $i$ at timestep $t$ for vehicle $m$ through the softmax function:

$$p_{i,m,t} = \text{softmax}(u_{i,m,t}) = \frac{e^{u_{i,m,t}}}{\sum_j e^{u_{j,m,t}}} \tag{17}$$

According to the probability $p_{i,m,t}$, we use sampling decoding in the training process and greedy decoding in the test process to choose the action. Greedy decoding means to select the best action with the maximum probability at each timestep, and sampling decoding means to sample several solutions and report the best.

*(3) Reward*

A reward function defines the goal of a reinforcement learning problem (Sutton and Barto, 1998). The reward function $R(r[1,M])$ is specified by

$$R(r[1,M]) = -\text{Cost}(r[1,M]) \tag{18}$$

### 3.4 Training Method

We parameterize the stochastic policy with parameters $\theta$, which is the vector of all trainable variables used in the encoder and decoder framework. To train the network, we use well-known policy gradient approaches. Policy gradient methods iteratively use an estimated gradient of the expected return to update the policy parameters. We optimize the parameter by the reinforce gradient estimator (Williams, 1992) with baseline $R(r^{BL}[1,M])$:

$$\nabla_\theta L(\theta|s) = -E_{r \sim p_\theta(\cdot|s)}[(R(r[1,M]) - R(r^{BL}[1,M]))\nabla_\theta \log p_\theta(r[1,M]|s)] \tag{19}$$

In Eq. (19), $R(r[1,M])$ is the cost of a solution from a deterministic sample decoding of the model according to the probability distribution $p_{i,m,t}$, which obtains a solution through sampling. $R(r^{BL}[1,M])$ is the cost of a solution from a deterministic greedy decoding of the baseline model. The baseline is used to estimate the difficulty of the problem instance $s$ and to eliminate the variance in the training process, such that it can relate to the cost to estimate the advantage of the solution selected by the model (Mnih et al.,2015, Kool et al., 2018). In addition, the baseline is stabilized by freezing the greedy rollout policy $p_{\theta^{BL}}$. Every epoch, we compare the current training model with the baseline model and replace the parameters $p_{\theta^{BL}}$ only if the improvement is significant in terms of a paired t-test (= 5%). Furthermore, we use the Adam optimizer to train the parameters by minimizing $\nabla_\theta L(\theta|s)$.

The training steps of the MAAM are illustrated in Algorithm 1.

*Algorithm 1*

---

**Input**  Generated a pool of problem instances $s \in \mathcal{S}$
**Output**  Solution $r[1, M]$
**Procedure**  MAAM training process
1: Initialize parameters $\theta$, $\theta^{BL}$
2: Compute the customer embedding $h_i^{(\lambda)}$ and aggregated embedding $\bar{h}^{(N)}$ by Eqs. (3)-(10) in the Encoder layer
3: **for each epoch do:**
4:   **for each batch do:**
5:     $t = 0$
6:     **for each vehicle** $m$ $(1 \leq m \leq M)$ **do:**
7:       Compute multiple vehicles context embedding $h_t^{(c)}$ by Eq. (12)
8:       Compute a new context embedding $h_t^{(c)'}$ by Eq. (13)
9:       Compute the output probability vector $p_{m,k,t}$ by Eqs. (14)-(17)
10:      Agent $m$ chooses action $r[m][t]$ using sample decoding
11:      Update the state
12:      $t = t + 1$
13:     **until** all of the customers have been served
14:     Compute reward $R$ for solution $r[1, M]$
15:     Compute reward $R^{BL}$ for solution $r^{BL}[1, M]$ using greedy decoding
16:     Compute reinforce gradient estimator with baseline $R^{BL}$ by Eq. (18) and update the parameters through the Adam optimizer
17:     If $PairedTTest\ (p_\theta, p_{\theta^{BL}}) < 5\%$:
18:       $\theta^{BL} = \theta$
19:   **End**
20: **End Procedure**

---

## 4 CASE STUDY

### 4.1 Experiment Setting

*(1) Instances with 20 customers*

We assume that the customers' locations, demands and time windows are randomly generated from a uniform distribution. Specifically, the depot location and twenty customers are randomly generated in the square $[0,10] \times [0,10]$. Such simulation settings on the Euclidean plane can be utilized in unmanned aerial vehicle delivery. The vehicle capacity is set to 60. The time window is randomly generated from [0,10]. Early and late penalty coefficients $\alpha_i$, $\beta_i$ are randomly generated from [0,0.2] and [0,1] separately. Each customer's demand is randomly generated from [0,10] for two vehicles and [0,15] for three vehicles. We evaluate our model on 1000 instances. It is worth mention that the total demands are controlled to less than the total capacity of all vehicles to ensure a feasible solution.

　　*(2) Instances with 50 customers*

In the medium-scale case, we randomly generate 50 customers. The vehicle capacity is set to 150. The time window is randomly generated from [0,20]. Each customer demand is randomly generated from [0,10] for two vehicles, [0,15] for three vehicles, [0,20] for four vehicles and [0,25] for five vehicles. Other parameter settings are similar to small-scale instances.

　　*(3) Instances with 100 customers*

In this case, we randomly generate 100 customers. The vehicle capacity is set to 300. The time window is randomly generated from [0,40]. Other parameter settings (including the vehicle numbers and customer demands) are similar to those in medium-scale instances.

　　*(4) Instances with 150 customers*

In this case, we randomly generate 150 customers and utilize 5 vehicles to serve them. The vehicle capacity is set as 180. The time window with length 20 is randomly generated from [0,60]. Early and late penalty coefficients $\alpha_i$, $\beta_i$ are set as 0.1 and 0.5 separately. Each customer's demand is randomly generated from [0,10].

**4.2 Benchmarks**

We directly compare the MAAM with two classical heuristic methods and commonly-used baseline Google OR-Tools.

　　*(1) Genetic algorithm*

The genetic algorithm (GA) is a prevalent method for solving both the constrained and unconstrained optimization problems, based on a natural selection process similar to biological evolution. GA repeatedly modifies a population of individual solutions. At each step, GA selects individuals at random from the current population to be parents and uses them to produce the children for the next generation. Over successive generations, the population evolves toward an optimal solution (Ombuki et al., 2006).

We adopt two sets of parameters: $GA^1$ with population size 100 and maximum iteration number 300; $GA^2$ with population size 300 and maximum iteration number 1000. Additionally, we set the crossover rate to 0.80 and the mutation rate to 0.05. This process is repeated until the maximum iteration is obtained or the algorithm's solutions remain unchanged for five iterations.

*(2) Iterated local search algorithm*

To provide the best possible result, iterative local search algorithms (ILS) move iteratively from one possible solution to a neighboring solution and so on until the best possible set of results is achieved. The algorithm continues to select solutions and their neighbors until there are no more improved configurations in the neighborhood, thus falling into a locally optimal solution. Furthermore, we make use of iterated local search to curb the tendency of falling into locally optimal points. However, it is impossible to quickly traverse all solutions in the neighborhood in consideration of the computational complexity, and thus, we set the termination criterion as a predetermined maximum iteration number. In this sense, the algorithm presents the best possible results within a stipulated number of iterations (Lourenço et al., 2003, Ibaraki et al., 2008).

We adopt two sets of parameters: $ILS^1$ with maximum iteration number 100, and $ILS^2$ with maximum iteration number 500. This process is repeated until the maximum iteration is obtained or the algorithm's solutions remain unchanged for five iterations.

*(3) Google OR-Tools*

The Google OR-Tools is an open source software suite for optimization, which is a widely used solver for the vehicle routing problems and constrained optimization. In the experiments, we implement the baselines using a parallel computation structure with 10 processes simultaneously.

*(4) Multi-agent attention model*

We trained the model for 100 epochs with randomly generated data under the learning rate of $10^{-4}$. In every epoch, 1,280,000 instances for small-scale instances and 640,000 instances for medium-scale and large-scale instances were processed. The batch size was set to 512 for small-scale instances and medium-scale instances and 256 for large-scale instances. Our experiments were performed on a computing platform as follows: NVIDIA Quadro P5000 with 16 GB memory, Intel(R) Xeon(R) CPU E5-2673 v3 @2.40 GHz with 256 GB RAM. The RL model is realized using Pytorch 1.1.0. We set the dimension of the initial customer embedding layer as 128, the number of layers *A* as 3, and the number of attention heads *Z* as 8 in the encoder. The following parameter sensitivity analysis demonstrates that the parameter setting is a good trade-off between the quality of the results and the computational complexity.

**4.3 Results**

Table 2 shows the total cost of each method under four different scale testing scenarios. The following conclusions can be drawn from the results:

i.  As shown in Table 2(a), all of the methods achieve similar results on small-scale problems. However, GAs perform much worse in comparison to other algorithms when the problem size grows. The OR-Tools is second only to our method. ILSs perform better than GAs, and $ILS^2$ is better than $ILS^1$ because we allow more iterations for the former.

ii. Our proposed model achieves the best performances compared with other baselines in terms of both the solution quality and the computation efficiency under most of scenarios. In fact, ILS fails to locate even sub-optimal solutions on large-scale problems, whereas our model can provide high-quality solutions within only a few seconds. Unlike most classical heuristic methods, it is robust to changes in the predefined conditions; for example, when a customer changes its demand value or relocates to a different position, it can automatically adapt the solution. In addition, according to comparison on the experiments with 150 customers, our model performs better than OR-Tools on the cases with total cost and only time window cost, while OR-Tools performs a bit better on the case with only travel cost. This is probably because MAAM is more suitable to deal with nonlinear constraint problem as a kind of neural network method.

iii. The computation time of our framework stays almost the same when the problem size increases, which is much faster than the benchmarks. In contrast, the run time for the heuristics methods and OR-Tools increases exponentially with the number of customers due to the NP-hard essence of the problem itself. Classical heuristic methods must re-solve the problem from beginning each time, while our MAAM framework can utilize some well-trained information to largely cut down the computational effort. This observation proves the superiority of our method.

Table 2. Performance comparison for MVRPSTW

*(a) Comparison on instances with 20 customers*

| Method | #Vehicle | Cost | Time | #Vehicle | Cost | Time |
|---|---|---|---|---|---|---|
| $GA^1$ | | 58.7 | 5(min) | | 67.0 | 4(min) |
| $GA^2$ | | 56.5 | 18(min) | | 66.8 | 14(min) |
| $ILS^1$ | 2 | 58.8 | 3(min) | 3 | 65.9 | 3(min) |
| $ILS^2$ | | 57.3 | 12(min) | | 64.8 | 10(min) |
| OR-Tools | | 55.8 | 30(sec) | | **63.8** | 23(sec) |
| **MAAM** | | **55.6** | **1(sec)** | | 64.3 | **1(sec)** |

*(b) Comparison on instances with 50 customers*

| Method | #Vehicle | Cost | Time | #Vehicle | Cost | Time |
|---|---|---|---|---|---|---|
| $GA^1$ | | 148.4 | 23(min) | | 134.5 | 22(min) |
| $GA^2$ | | 115.7 | 1.5(hrs) | | 118.0 | 1.3(hrs) |
| $ILS^1$ | 2 | 108.0 | 14(min) | 3 | 102.2 | 13(min) |
| $ILS^2$ | | 102.4 | 1.1(hrs) | | 100.3 | 1(hr) |
| OR-Tools | | 93.0 | 13(min) | | 94.8 | 9(min) |
| **MAAM** | | **87.6** | **2(sec)** | | **93.5** | **2(sec)** |
| GA[1] | | 156.3 | 18(min) | | 157.8 | 15(min) |
| GA[2] | | 127.6 | 1(hrs) | | 129.6 | 56(min) |
| ILS[1] | 4 | 113.7 | 7(min) | 5 | 121.6 | 6(min) |
| ILS[2] | | 112.1 | 45(min) | | 120.3 | 41(min) |
| OR-Tools | | 103.7 | 6(min) | | 112.4 | 5(min) |
| **MAAM** | | **101.9** | **2(sec)** | | **112.1** | **2(sec)** |

*(c) Comparison on instances with 100 customers*

| Method | #Vehicle | Cost | Time | #Vehicle | Cost | Time |
|---|---|---|---|---|---|---|
| GA | | 278.7 | 6.2(hrs) | | 267.5 | 5.4(hrs) |
| ILS[1] | 2 | 181.6 | 1.1(hrs) | 3 | 161.6 | 56(min) |
| OR-Tools | | 146.4 | 1.8(hrs) | | 139.8 | 1.5(hrs) |
| **MAAM** | | **131.5** | **5(sec)** | | **132.3** | **5(sec)** |
| GA | | 263.5 | 4.5(hrs) | | 281.4 | 4.1(hrs) |
| ILS[1] | 4 | 177.8 | 35(min) | 5 | 187.9 | 34(min) |
| OR-Tools | | 158.7 | 1.1(hrs) | | 163.6 | 52(min) |
| **MAAM** | | **139.4** | **4(sec)** | | **146.5** | **4(sec)** |

*(d) Comparison on instances with 150 customers*

| | Total Cost | | Only travel cost | | Only time window cost | |
|---|---|---|---|---|---|---|
| | Cost | Time | Cost | Time | Cost | Time |
| ILS[1] | 243.6 | 2.1(hrs) | 145.7 | 32(min) | 1.78 | 2.1(hrs) |
| OR-Tools | 200.8 | 2.9(hrs) | **122.9** | 4(min) | 1.30 | 3.1(hrs) |
| **MAAM** | **197.3** | **6(sec)** | 125.7 | **3(sec)** | **1.27** | **6(sec)** |

To show the training process in detail, we visualize the reward during the training process in Figure 4. "20C-2V" represents the MVRPSTW problem with 20 customers and 2 vehicles as a simplified expression. We can find that the curve converges gradually with increasing epochs. In addition, with the increase in the number of customers, the training loss becomes more unstable. Furthermore, the trained model can already present a fine result after training for 20 epochs, which means that we can sharply reduce the training time if the requirements for the solution quality are not overly demanding.

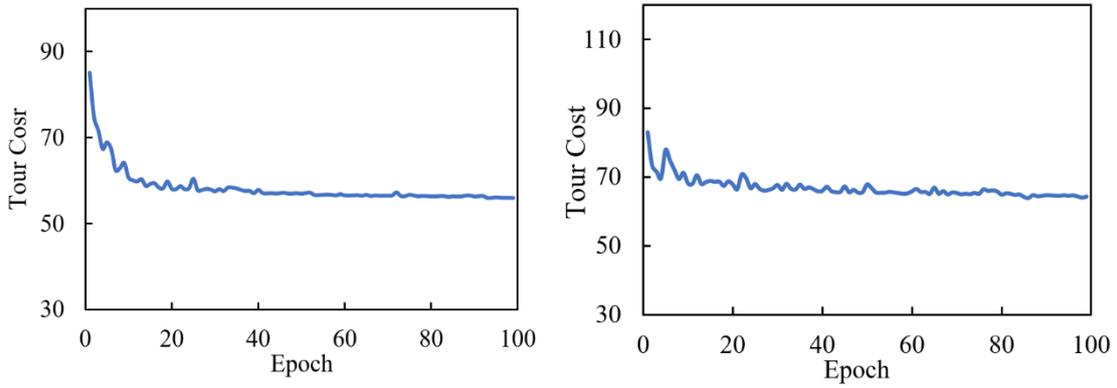

(a) Tour cost for 20C-2V and 20C-3V

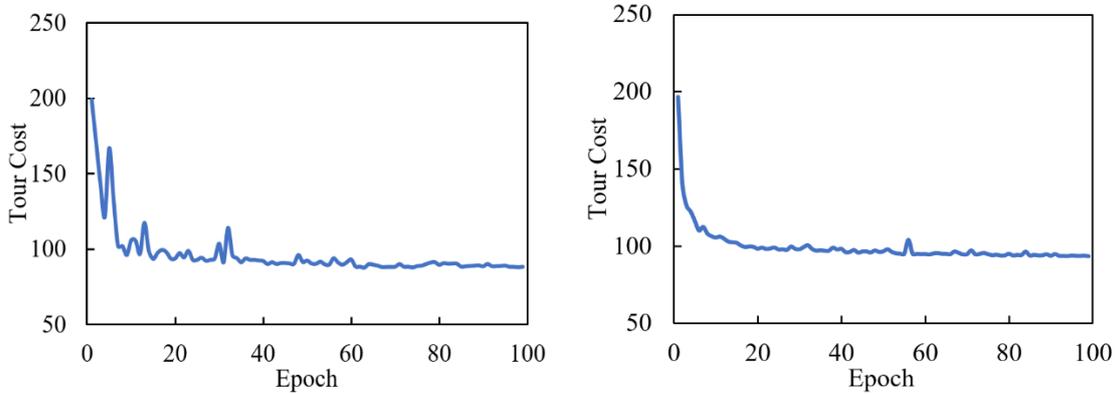

(b) Tour cost for 50C-2V and 50C-3V

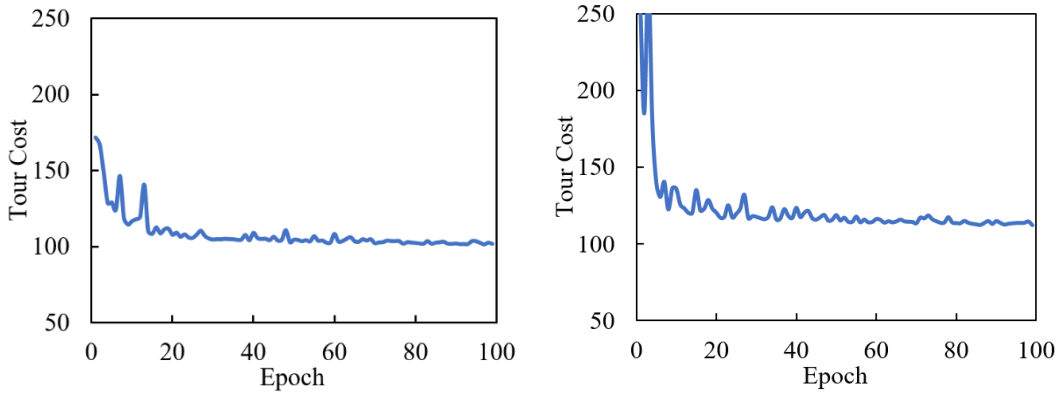

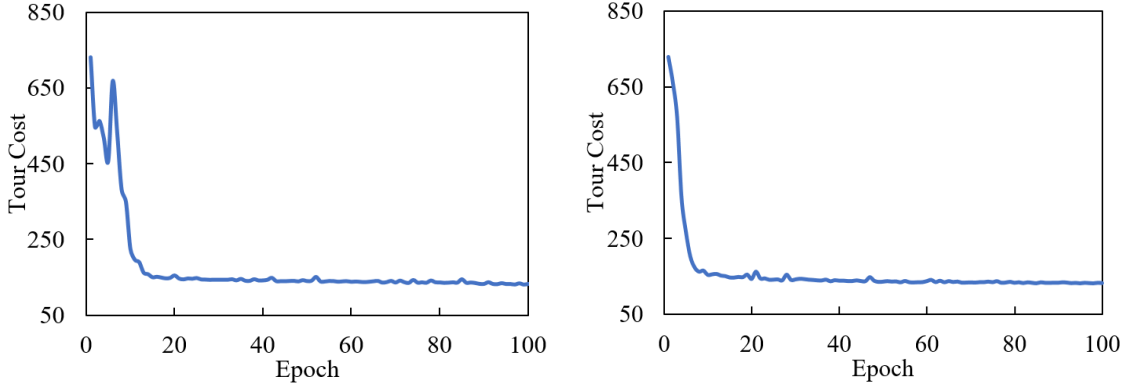

(c) Tour cost for 50C-4V and 50C-5V

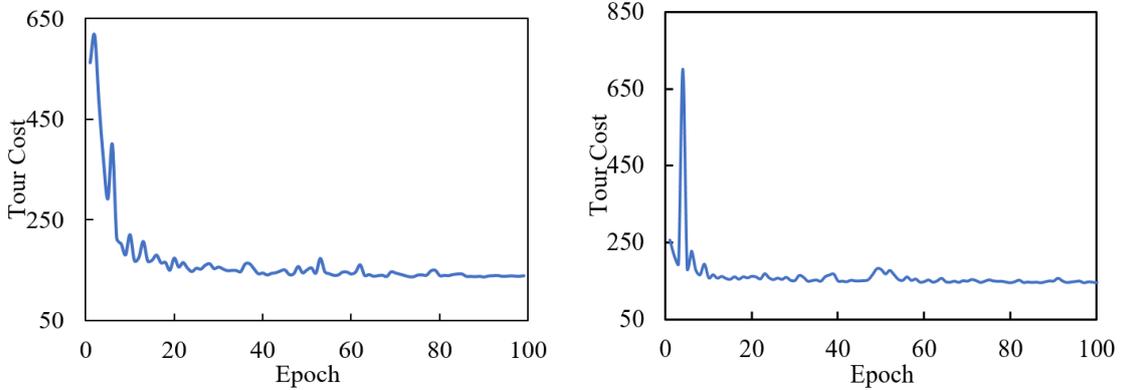

(d) Tour cost for 100C-2V and 100C-3V

(e) Tour cost for 100C-4V and 100C-5V

Figure 4: Tour cost curves in the training process

**4.4 Sensitivity Analyses**

In this section, we analyze the parameter sensitivities in the proposed model, which could greatly influence the solution quality of our MAAM. Three parameters are investigated in this section: the dimension of the initial customer embedding in the encoder framework, the number of encoder layers, and the number of attention heads.

First, we retrain the network by setting the embedding dimension of the customer as 64, 128, and 256, and we assess the corresponding performance. The reward curves are depicted in Figure 5(a). It is obvious that the model tends to converge faster with increased embedding dimensions. This finding arises because some useful information will be neglected in low dimensional space, which leads to deteriorated algorithm output. The training times with the dimensions of 256 and 128 are 670 seconds and 590 seconds, respectively.

Similarly, we test the sensitivity of the encoder layers $A$ in [2,3,4], and the reward curves are demonstrated in Figure 5(b). This finding shows that an overly shallow structure (i.e., two layers) makes it difficult to capture the information among the customers, but a deeper neural network does not always give a better result.

Finally, we evaluate the influences of the multi-head attention mechanism with different numbers of attention heads $Z$. The variation curves of the rewards with the attention head number are plotted in Figure 5(c). There is an obvious improvement when adding the attention head from two to four. This finding could be caused by the fact that the self-attention mechanism can effectively represent the probability of a relationship between the terms of the customer embedding and can find a new representation for each of the terms in the sequence for the decisions. It is worthwhile to note that the profit becomes inconspicuous with more attention heads, whereas the computational time rises vastly. The comparison results demonstrate the effectiveness of applying multi-head self-attention to extract the features.

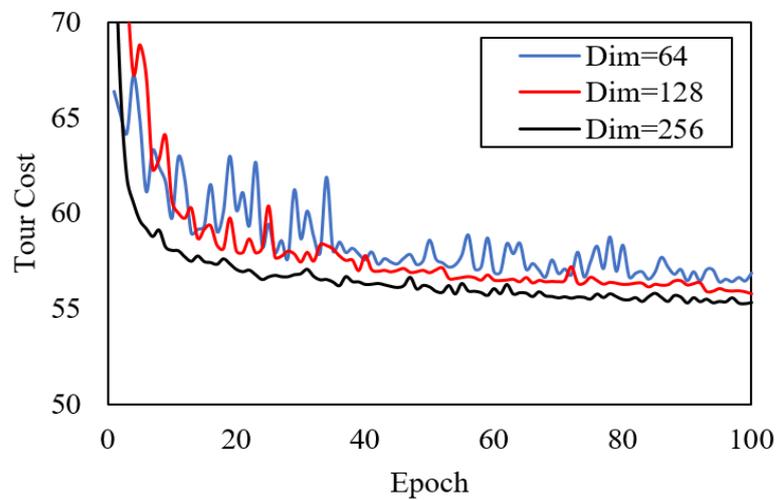

(a) Dimension of initial customer embedding in the encoder framework

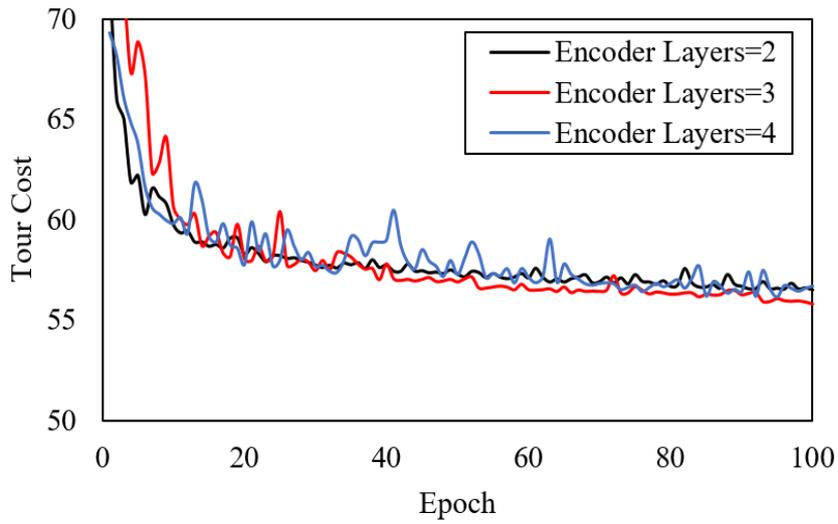

(b) Number of encoder layers

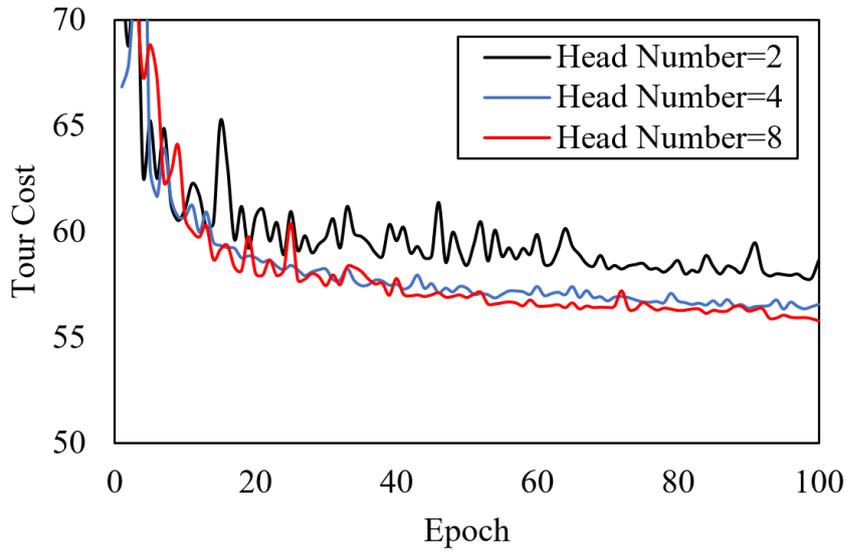

(c) Number of attention heads

Figure 5. Sensitivity analyses in MAAM parameters

**4.5 Method Robustness**

In the real world, the number of customer requests and the vehicle capacity always exhibit fluctuations over time, which requires the model to be able to address such stochasticity. To resolve this problem, we only need to add some virtual customers in the original stage. To be more specific, we have a well-trained model for 100 customers, but during a certain day there are only 98 requests, and in this case, we add two virtual customers with zero demand and the same location and time window as any existing customer to satisfy the number of constraints. To verify the robustness of our model, we designed two experiments in this section. In the first experiment, we use the well-trained 50C-2V model and evaluate its performance on the 40C-2V, 48C-2V, 46C-2V, 44C-2V, 42C-2V, and 40C-2V cases. Furthermore, we use the well-trained 100C-2V model and evaluate its performance on the problem with different numbers of customers, varying from 90 to 98. As shown in Table 3, our method consistently outperforms OR-Tools and ILS[1] in terms of the solution quality and computation time.

Table 3. Performance of well-trained models on variable customer numbers

*(a) Well-trained model with 50 customers*

| Method | #Customer | Cost | Time | #Customer | Cost | Time |
| --- | --- | --- | --- | --- | --- | --- |
| ILS[1] |  | 108.0 | 14(min) |  | 105.0 | 14(min) |
| OR-Tools | 50 | 93.0 | 13(min) | 48 | 91.4 | 12(min) |
| **MAAM** |  | **87.6** | **2(sec)** |  | **85.6** | **2(sec)** |
| ILS[1] |  | 99.8 | 13(min) |  | 96.0 | 13(min) |
| OR-Tools | 46 | 87.4 | 10(min) | 44 | 84.6 | 9(min) |
| **MAAM** |  | **83.8** | **2(sec)** |  | **82.0** | **2(sec)** |
| ILS[1] |  | 92.0 | 12(min) |  | 88.9 | 11(min) |
| OR-Tools | 42 | 81.2 | 9(min) | 40 | 79.8 | 7(min) |
| **MAAM** |  | **80.5** | **2(sec)** |  | **79.1** | **2(sec)** |

*(b) Well-trained model with 100 vehicles*

| Method | #Customer | Cost | Time | #Customer | Cost | Time |
|---|---|---|---|---|---|---|
| ILS[1] |  | 181.6 | 1.1(hrs) |  | 179.9 | 1.1(hrs) |
| OR-Tools | 100 | 146.4 | 1.8(hrs) | 98 | 144.5 | 1.7(hrs) |
| **MAAM** |  | **131.5** | **5(sec)** |  | **129.8** | **5(sec)** |
| ILS[1] |  | 172.4 | 1.1(hrs) |  | 168.4 | 1(hrs) |
| OR-Tools | 96 | 142.8 | 1.7(hrs) | 94 | 139.4 | 1.6(hrs) |
| **MAAM** |  | **127.5** | **5(sec)** |  | **125.4** | **5(sec)** |
| ILS[1] |  | 163.0 | 58(min) |  | 159.8 | 56(min) |
| OR-Tools | 92 | 137.2 | 1.6(hrs) | 90 | 135.1 | 1.5(hrs) |
| **MAAM** |  | **124.1** | **4(sec)** |  | **122.5** | **4(sec)** |

In the second experiment, the generality of our model is tested when the capacity of the vehicles varies. The vehicle capacity is fixed in the training process, and as a result, we must adjust the demands of the customers in inverse proportion to the changes in the capacity, e.g., we multiply all demands by 0.5 if the vehicle capacity is doubled. We use the well-trained models for 50C-2V to generate a solution for the same problem with different vehicle capacities that range from 120 to 180. Then, we use the models trained for 100C-2V to solve the problem with different vehicle capacities that range from 270 to 330. Table 4 shows that the well-trained method receives good results compared with an iterated located search.

Overall, the comparison results in the two experiments indicate that when the problems are close in terms of the number of customers and vehicle capacity, our well-trained model can still produce significantly better vehicle routes. This finding demonstrates that our model is robust to variations in the problem instances, e.g., when several customers cancel their demands or the vehicle capacity is adjusted.

Table 4. Performance of well-trained models on variable vehicle capacities

*(a) Well-trained model with 50 vehicles*

| Method | Capacity | Cost | Time | Capacity | Cost | Time |
|---|---|---|---|---|---|---|
| ILS[1] |  | 114.3 | 9(min) |  | 107.2 | 16(min) |
| OR-Tools | 120 | 94.7 | 11(min) | 180 | 92.1 | 14(min) |
| **MAAM** |  | **87.8** | **2(sec)** |  | **87.5** | **2(sec)** |

*(b) Well-trained model with 100 vehicles*

| Method | Capacity | Cost | Time | Capacity | Cost | Time |
|---|---|---|---|---|---|---|
| ILS[1] |  | 189.8 | 52(min) |  | 181.1 | 1.2(hrs) |
| OR-Tools | 270 | 146.8 | 1.5(hrs) | 330 | 145.7 | 1.8(hrs) |
| **MAAM** |  | **132.3** | **5(sec)** |  | **132.0** | **5(sec)** |

## 5 CONCLUSIONS

In this paper, we propose a novel reinforcement learning algorithm called the multi-agent attention model (MAAM) to solve MVRPSTW. According to the results of simulation experiments on four synthetic networks with different scales, our proposed MAAM consistently outperforms Google OR-Tools and traditional methods with negligible computation time, which suggests that there would be successful adoption of deep reinforcement learning (DRL) for VRPs with complicated practical constraints. Moreover, unlike many time-consuming classical heuristics, our proposed method has superior performances in both the solution quality and efficiency. In addition, we find our well-trained model has a certain level of robustness to solve problems with fluctuations in the customer numbers and vehicle capacities, and the robust performance extends the applicability of the model to handle more realistic cases.

In future research, it will be an important topic to utilize a machine-learning-based method to solve more combinatorial optimization problems of practical importance, e.g., VRPs with multiple depots, multiple periods, heterogeneous vehicle fleets, and so on. Extending the methodology to solve very large-scale problems with thousands of vehicles and customer requests is also of great interest. A more challenging task is to generalize the learning framework into online problems, i.e., to address the possibilities of real-time requests as well as stochastic traffic conditions.

**ACKNOWLEDGEMENTS**

The research is supported by grants from Science & Technology Program of Beijing (Z191100002519008), National Natural Science Foundation of China (71871126), and Tsinghua-Toyota Joint Research Institution.